\documentclass[runningheads]{llncs}
\usepackage{graphicx}
\usepackage{amsmath}
\usepackage{amssymb}
\usepackage{amsfonts}
\usepackage{bbold}

\usepackage{url}
\usepackage{microtype}      % microtypography
\usepackage{multirow}
\usepackage{subfigure}
\usepackage{booktabs}       % professional-quality tables
\usepackage{algorithm}
\usepackage{algpseudocode}
\usepackage{mathtools}
\usepackage{xcolor}
\usepackage{wrapfig}

\DeclareMathOperator{\Tr}{Tr}
\DeclareMathOperator{\Diag}{Diag}
\begin{document}

\title{Multi-Modal Unsupervised Pre-Training for Surgical Operating Room Workflow Analysis}
%Unsupervised Representation Learning via clustering for OR understanding.
%Unsupervised Representation Learning via clustering for OR workflow analysis
%Unsupervised Representation Learning via Clustering for Surgical Video Activity Recognition and Semantic Segmentation

\author{Muhammad Abdullah Jamal and Omid Mohareri}
\institute{Intuitive Surgical Inc., Sunnyvale, CA}%\\ \{abdullah.jamal, omid.mohareri\}@intusurg.com 
% \author{}
% \institute{}

\maketitle

\begin{abstract}
% One of the main bottlenecks for data-driven methods for OR workflow analysis is manually annotating and labelling of a data. Can we reduce the annotation time, exploit the large number of unlabeled data, and eventually fine-tune the model with less labeled data? To answer this question, we propose a novel way to fuse both the intensity and the depth map of a single frame or image, and train a clustering-based unsupervised approach to extract discriminative representations. Inspired by recent clustering-based approach, we first unwrap the frames or images into intensity and depth and then train a “SwAV” model~\cite{swav} for learning unsupervised representations. Instead of producing different augmentations (or “views”) of the same image, we leverage the intensity and the depth maps of the frame and treat them as two different views. We compared our method with other state of the art methods and results show the superior performance of our approach on surgical video activity recognition and semantic segmentation.

Data-driven approaches to assist operating room (OR) workflow analysis depend on large curated datasets that are time consuming and expensive to collect. On the other hand, we see a recent paradigm shift from supervised learning to self-supervised and/or unsupervised learning approaches that can learn representations from unlabeled datasets. In this paper, we leverage the unlabeled data captured in robotic surgery ORs and propose a novel way to fuse  the multi-modal data for a single video frame or image. Instead of producing different augmentations (or “views”) of the same image or video frame which is a common practice in self-supervised learning, we treat the multi-modal data as  different views to train the model in an unsupervised manner via clustering. We compared our method with other state of the art methods and results show the superior performance of our approach on surgical video activity recognition and semantic segmentation.

%Inspired by recent clustering-based approach, we first unwrap the frames or images into intensity and depth and then train a “SwAV” model~\cite{swav} for learning unsupervised representations. 
%both the intensity and the depth map.
%the intensity and the depth map

\end{abstract}

\section{Introduction}\label{sec:intro}
Robotic Surgery has allowed surgeons to perform complex surgeries with more precision and with potential benefits such as shorter hospitalization, fast recoveries, less blood lost etc. However, it might not be easy to adopt due to cost, training, integration with the existing systems, and OR workflow complexities~\cite{Catchpole2015SafetyEA} that can eventually lead to human errors in OR.

Recently, data-driven based workflow analysis has been proposed to help identify and mitigate these errors. Sharghi et al.~\cite{aidean} proposes a new dataset that includes 400 full length videos captured from several surgical cases. They used 4 Time of Flight (ToF) cameras to generate a multi-view dataset. Moreover, they also propose a framework to automatically detect surgical activities inside OR. Li et al.~\cite{Li} proposes a new dataset and framework for semantic segmentation in OR. Schmidt et al.~\cite{Schmidt} leverages multi-view information and proposes a new architecture for surgical activity detection. However, all these approaches require expert people to manually annotate the data which is expensive and time consuming. Multi-model data/sensing can provide richer information about the scene that can benefit higher performing visual perception tasks like action recognition or object detection. Cameras such as RGB-D, Time of Flight (ToF) have been used in OR to capture the depth information, 3D point clouds, and intensity maps etc. For example~\cite{mvor} uses RGB-D cameras to capture data for 2D/3D pose estimation in hybrid OR. To the best of our knowledge, very little or no work has been done to leverage multi-modalities for surgical OR understanding under unsupervised setting. 
%Kadkhodamohammadi et al.~\cite{Kadkhodamohammadi} also proposes a novel solution to capture open surgery procedures and develop a CNN-LSTM based approach to automatically segment the videos into pre-defined surgical phases
In this paper, we are focusing on leveraging unlabeled multi-modal image and video data collected in the OR. We propose an unsupervised representation learning approach based on clustering that can help alleviate the annotation time in data-driven approaches. Recently, unsupervised or self-supervised methods have been proposed which have significantly lessen the performance gap with the supervised learning~\cite{simclr,moco,cpc,pirl}. These methods mostly rely on contrastive loss and set of transformations. They generate different augmentations or views of an image and then directly compare the features using contrastive loss so that they can push away representations from different images and pull together the representations from different views of the same image. But this may not be a scalable approach as it could require computation of all pairwise comparisons on large datasets, and often need larger memory banks~\cite{moco} or larger batch size~\cite{simclr}. On the other hand, clustering-based approaches group the semantically similar features instead of individual images. We are also interested in semantically similar group of features that can provide better representations especially using multi-modal data for image and video understanding tasks in surgical operating rooms. To address the privacy concerns, and preserve the anonymity of people in OR, we limit ourselves to intensity and the depth maps captured from the Time of Flight (ToF) cameras. 

To this end, we propose a novel way to fuse the intensity and the depth map of a single video frame or an image to learn representations. Inspired by~\cite{swav}, we learn the representations in an unsupervised manner via clustering by considering the intensity and the depth map as two different 'views' instead of producing different augmentations ('views') of the same video frame or image. The features of the two views are mapped to the set of learnable prototypes to compute cluster assignments or codes which are subsequently predicted from the features. If the features capture the same information, then it should be possible to predict the code of one view from the feature of the other view. We validate the efficacy of our approach by evaluating on surgical video activity recognition and semantic segmentation in OR. In particular, we achieve superior performance as compared to self-supervised approaches designed especially for video action recognition under various data regime. We also show that our approach achieve better results on semantic segmentation task under low-data regime as compared to clustering based self-supervised approaches, namely Deep Cluster~\cite{deepcluster}, and SELA~\cite{sela}.
\section{Related Work}
Our work is closely related to data-driven approaches for OR workflow analysis briefly reviewed in section~\ref{sec:intro}. In this section, we will further review some data-driven approaches for OR workflow analysis, and some self-supervised approaches.

\paragraph{\textbf{Data-driven approaches for OR workflow.}}
~\cite{Kadkhodamohammadi16,Kadkhodamohammadi17} use multi-view RGB-D data for clinician detection and human pose estimation in OR.~\cite{HPE_low} introduced a multi-scale super-resolution architecture for human pose estimation using low-resolution depth images.~\cite{face_mvor} compared several SOTA face detectors using MVOR dataset, and then propose a self-supervised approach which learns from unlabeled clinical data.

%~\cite{mvor} introduced Multi-View Operating Room (MVOR) dataset to address 2D and 3D human pose estimation in the OR

\paragraph{\textbf{Self-supervised Learning.}} 
Self-supervised approaches learn representations using unlabeled data by defining a pre-text task which provides the supervised signal. Earlier approaches use reconstruction loss with the autoencoders as pretext task to learn unsupervised features~\cite{denoising,reconstruction}. Masked-patch based prediction models~\cite{Bert,MaskedAE} which use autoencoders have also been proposed in the similar context. Recently, self-supervised learning paradigm has shifted towards the instance discrimination based contrastive learning~\cite{BYol,cpc}. It considers every image in the data as its own class and learns the representations by pulling together features of different augmentations of the same image. We have also discussed few contrastive learning based approaches in section~\ref{sec:intro}. Besides, there are several methods that learn visual features by grouping samples via clustering~\cite{sela,deepcluster}. Our work is closely related to the clustering-based approaches.~\cite{deepcluster} learns representations using k-means assignments which are used as pseudo-labels.~\cite{sela} uses optimal transport to solve the pseudo-labels assignment problem.~\cite{swav} contrasts cluster assignments instead of features which enforces the consistency between different augmentations of the image. Finally, several pretext tasks have also been proposed for video domain. It includes pace prediction~\cite{pace,playback}, frame and clip order prediction~\cite{COP,OPN}, and contrastive prediction~\cite{Feichtenhofer,cvrl}.

%There are other hand-crafted pretext tasks that include colorizaton~\cite{color}, jigsaw puzzle~\cite{jigsaw}, and predicting image rotation~\cite{rotnet}{Zhirong->byol}{pmlr-v15-coates11a->clustering}
\section{Method}
% We first introduce the problem statement and then explain how to fuse intensity and the depth map from OR data to learn unsupervised representations inspired by~\cite{swav}.
% \paragraph{Problem Statement.}
Our goal is to learn representations using unlabeled OR data via clustering by fusing intensity and the depth maps captured using Time of flight (ToF) cameras. Previous clustering-based unsupervised approaches~\cite{sela,deepcluster} work in offline manner where they first cluster the features of the whole dataset and then compute the cluster assignments for different views. This is not feasible in large-scale setting as it requires to compute features of the entire dataset multiple times. 

%Previous clustering-based approaches are offline in nature where they alternate between cluster-assignment in which the whole dataset is clustered and the prediction of the cluster-assignment (“codes”). Similar to SwAV, we want our approach to scale to any dataset size which is not feasible with the offline clustering approaches. Therefore, we limit ourselves to online learning 

% Inspired by recent clustering-based approach called SwAV~\cite{swav}, which contrasts different augmentations of the image by comparing cluster assignments. It is essentially a swapped prediction method where the cluster assignment of one augmented data is predicted using the feature of other augmented data of the same image. 
Inspired by recent clustering-based approach called SwAV~\cite{swav} which works in an online manner by computing cluster assignments using features within a batch, we also limit our selves to an online learning and computes the cluster assignments within a batch using the representations from our multi-modal data. But, unlike SwAV, we don't produce different augmentations (views) of the same image, we treat the intensity and the depth map as two different views of the same video frame or image.
%which enforces the consistency between codes obtained from different augmentations of an image, and contrasts them by comparing cluster assignments.
%solves a swapped prediction problem wherein the cluster assignment of one augmented data is predicted using the feature of other augmented data of the same image.
%They also treat the augmentations as "views" in their work.

\begin{figure*}
    \centering
    \includegraphics[width=1.0\textwidth]{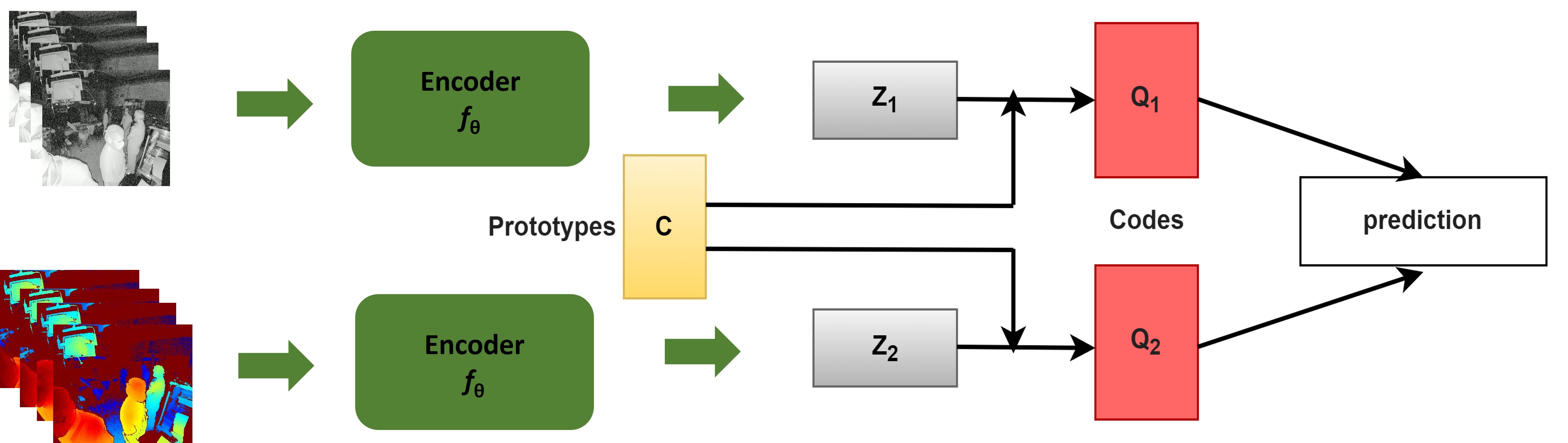}
    \caption{Our approach takes an intensity and the depth map, and then it extracts features \textbf{z$_{1}$} and \textbf{z$_{s}$} from the encoder $f_{\theta}$. Next, it computes the codes \textbf{q$_{1}$} and \textbf{q$_{2}$} by mapping these features to the set of \textbf{$K$} learnable prototypes \textbf{$\mathcal{C}$}. Finally, it predicts the code of one sample from the representation of the other sample. }
    \label{fig:approach}
\end{figure*}

\subsection{Fusion of Intensity and Depth Maps}
Given an intensity \textbf{x$_{1}$} and the depth map \textbf{x$_{2}$} of a single image or video frame, we first learn the representation through encoder $f$ which is parameterized by $\theta$ as \textbf{z$_{1}$} = $f_{\theta}$(\textbf{x$_{1}$}) and \textbf{z$_{2}$} = $f_{\theta}$(\textbf{x$_{2}$}) respectively. Then, we compute the codes \textbf{q$_{1}$} and \textbf{q$_{2}$} by mapping these representations to the set of \textbf{$K$} learnable prototypes $\{c_{1}, . . . , c_{K}\} \in$ \textbf{$\mathcal{C}$}. Finally, codes and representations are then used in the following loss function. Fig~\ref{fig:approach} illustrates the main idea.

\begin{equation}
\mathcal{L}(\textbf{z}_{1}, \textbf{z}_{2}) = l(\textbf{z}_{1}, \textbf{q}_{2}) + l(\textbf{z}_{2}, \textbf{q}_{1}) \label{loss}
\end{equation}

where $l(\textbf{z}, \textbf{q})$ represents the cross-entropy loss between the probability obtained by softmax on the dot product between \textbf{z} and \textbf{$\mathcal{C}$} and and the code \textbf{q} which is given as:

\begin{equation}
l(\textbf{z}_{1}, \textbf{q}_{2}) =  -\sum_{k} \textbf{q}_{2}^{(k)} \log \textbf{p}_{1}^{(k)}, \text{where}~\textbf{p}_{1}^{(k)} = \frac{\exp({\textbf{z}_{1} \cdot c_{k} /\tau })}{\sum_{k'}\exp({\textbf{z}_{1} \cdot c_{k'} /\tau })}
\label{cress-ent}
\end{equation}

where $\tau$ is a temperature hyperparameter. The intuition here is that if the representations \textbf{z$_{1}$} and \textbf{z$_{2}$} share the same information, then it should be possible to predict the code from the other representation. The total loss is calculated over all the possible pair of intensity and depth maps, which is then minimized to update the encoder $f_\theta$ and prototype \textbf{$\mathcal{C}$} which is implemented as a linear layer in the model. It is also straight forward to apply our fusion approach to learn representations for unlabeled video. Instead of using a single intensity and depth pair, we can sample a video clip, where each frame in the clip can be splitted into intensity and depth map. We will empirically show the superiority of our approach in the surgical video activity recognition in the section~\ref{sec:exp}.

\subsection{Estimating Code \textbf{q}}
The codes \textbf{q$_{1}$} and \textbf{q$_{2}$} are computed within a batch by mapping the features \textbf{z$_{1}$}, and \textbf{z$_{2}$}  to the prototypes \textbf{$\mathcal{C}$} in an online setting. Following~\cite{swav}, we ensure that all the instances in the batch should be equally partitioned into different clusters, thus avoiding the collapse of assigning them to a single prototype. Given feature vectors \textbf{Z} whose columns are \textbf{z$_{1}$}, ..., \textbf{z$_{B}$} which are mapped to prototypes \textbf{$\mathcal{C}$}. We are interested in optimizing this mapping or codes \textbf{Q} $=$ \textbf{q$_{1}$},..., \textbf{q$_{B}$}. We optimize using Optimal Transport Solver~\cite{OTP} as:

\begin{equation}\label{OTS}
    \max_{\textbf{Q} \in \mathcal{Q}} \Tr(\textbf{Q}^T\textbf{C}^T\textbf{Z}) + \epsilon \mathcal{H}(\textbf{Q})
\end{equation}

where $\mathcal{H}$(\textbf{Q}) corresponds to the entropy and $\epsilon$ is the hyperparameter that avoids the collapsing. Similar to~\cite{swav}, we restricted ourselves to mini-batch settings, and adapt their solution for transportation polytope as:

\begin{equation}\label{TP}
    \mathcal{Q} = \bigg\{\textbf{Q} \in \mathcal{R}^{K \times B} \mid \textbf{Q} \mathbb{1}_{B} = \frac{1}{\textbf{K}} \mathbb{1}_{\textbf{K}}, \textbf{Q}^T \mathbb{1}_{\textbf{K}} = \frac{1}{\textbf{B}} \mathbb{1}_{\textbf{K}} \bigg\}
\end{equation}

where $\mathbb{1}_{\textbf{K}}$ corresponds to the vector of ones with dimension \textbf{K}. The soft assignments \textbf{Q*} found for equation~\ref{OTS} are estimated using iterative Sinkhorn-Knopp algorithm~\cite{sinkhorn}. It can be written as:

\begin{equation}
    \textbf{Q*} = \Diag(\lambda)\exp \bigg({\frac{\textbf{C}^T\textbf{Z}}{\epsilon}}\bigg) \Diag(\mu)
\end{equation}

where $\lambda$ and $\mu$ correspond to renormalization vectors. We also use a queue to store the features from the previous iteration because the batch size is usually smaller as compared to the number of prototypes.
\section{Experiments}\label{sec:exp}
The goal of our pre-training approach is to provide better initialization for downstream tasks such as surgical activity recognition, semantic segmentation etc., and it is applicable to both video and image domain. We evaluated our learned representation on two different tasks i.e., surgical video activity recognition and semantic segmentation under low-data regime. We use mean average precision (mAP) and mean intersection over union (IOU) as evaluation metrics. 

\begin{figure*}
    \centering
    \subfigure{
    \includegraphics[width=0.6\textwidth]{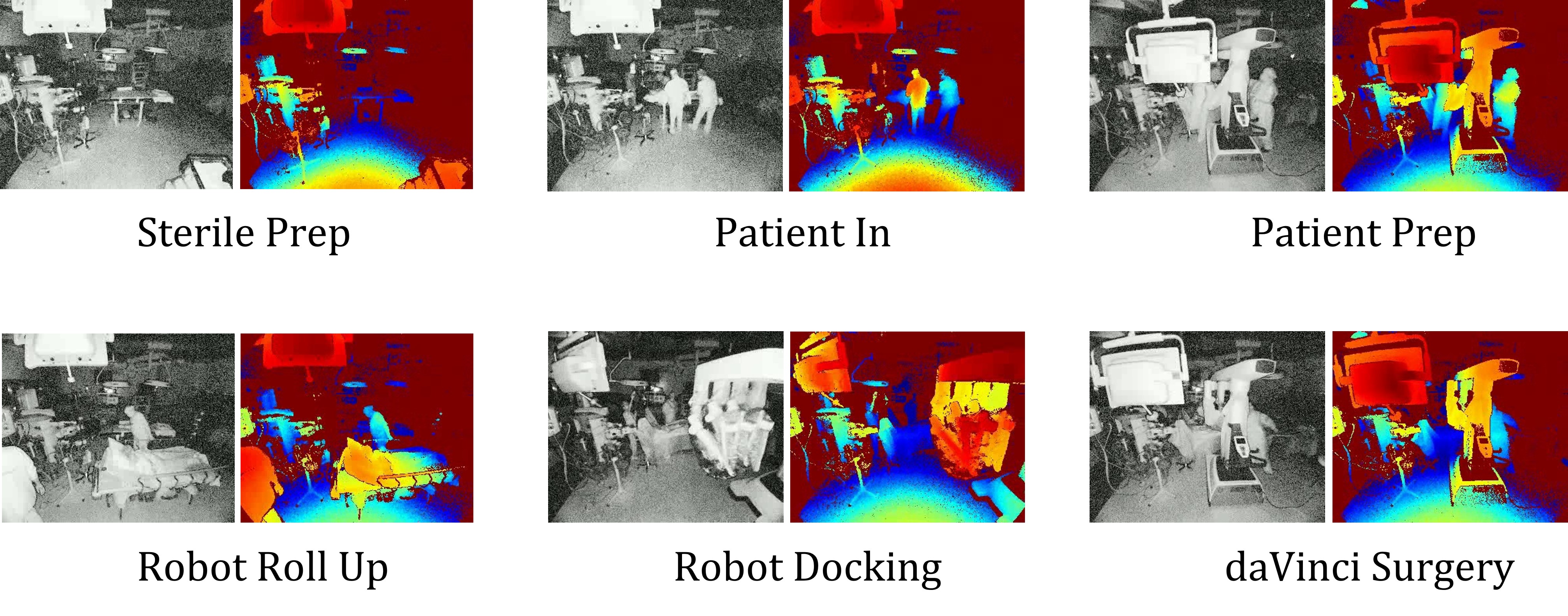}
    }
    \centering
    \subfigure{
    \includegraphics[width=0.33\textwidth]{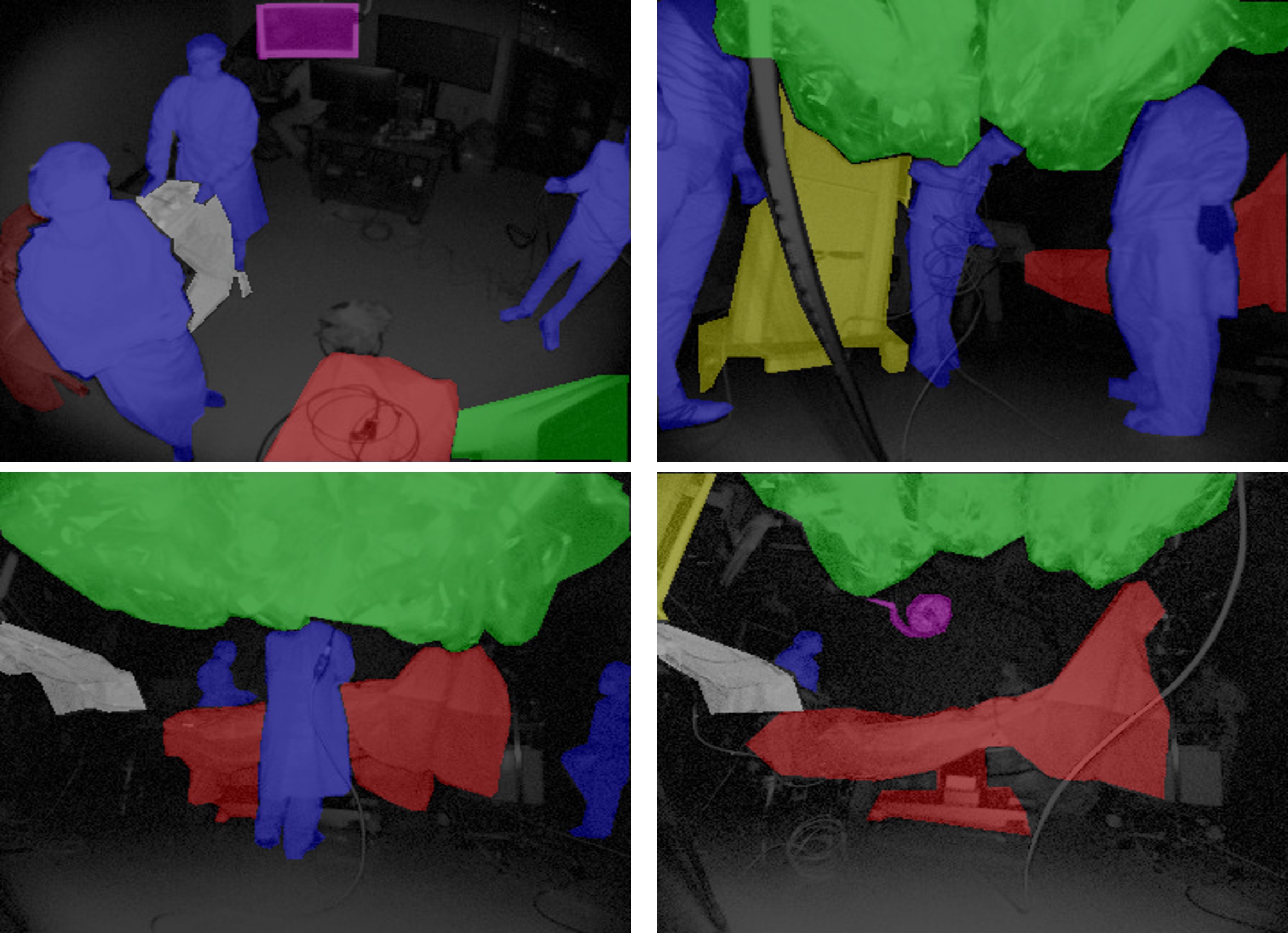}
    }
    \caption{a). OR Activity Dataset~\cite{aidean,Schmidt} with the intensity on the left side and the depth map on the right side for each activity. b). Semantic Segmentation Dataset~\cite{Li}. }
    \label{fig:activity}
    %\vspace{-5pt}
\end{figure*}

\subsection{{Datasets}}
We evaluated our approach on the following two datasets.

\paragraph{\textbf{Surgical Video Activity Recognition.}}
This dataset has been proposed by~\cite{aidean,Schmidt} that consists of 400 full-length videos captured from 103 surgical cases performed by the da Vinci Xi surgical system. The videos are captured from four Time of Flight (ToF) cameras placed on two vision carts inside the operating room. The videos are annotated by a trained user with 10 clinical activities. The intensity and the depth maps are extracted from the raw data from ToF camera. The dataset is splitted into 70\% training set and 30\% test set. During dataset preparation, we also make sure that all 4 videos belonging to the same case are either in train or test set. Please refer to left panel of the fig~\ref{fig:activity} for example activities.

\paragraph{\textbf{Semantic Segmentation.}}
~\cite{Li} has proposed densely annotated dataset for semantic segmentation task. It consists of 7980 images captured by four different Time of Flight (ToF) camera attached in the OR. We first split the dataset into training set (80\%) and testing set (20\%). From the training set, we create four different subsets of training data by varying the percentage of labels. The right panel of fig~\ref{fig:activity} shows few overlay images from the dataset.

%The dataset has been captured in a clinical scenarios, and videos are taken by the ToF cameras.

% \begin{figure*}
%     \centering
%     \includegraphics[width=0.8\textwidth]{Figs/ss_dataset.jpg}
%     \caption{Semantic Segmentation Dataset~\cite{Li}.}
%     \label{fig:ss}
%     %\vspace{-5pt}
% \end{figure*}

\subsection{Surgical Activity Recognition}
We use I3D~\cite{i3d} as a backbone architecture for this experiment, but our approach can be applied to any of the other SOTA models like TimeSFormer~\cite{Timesformer}, SlowFast~\cite{slowfast} etc. For all methods, we first train the I3D model on video clips similar to~\cite{aidean}, and then extract features for full videos from I3D to train Bi-GRU to detect surgical activities. Please refer to supplementary material for implementation details.

\paragraph{\textbf{Competing methods.}}
We compare our approach to the following competing baselines.
\begin{itemize}

\item \textbf{Baseline.} This is the baseline that either trains the I3D model on video clips from the scratch or trains the model pre-trained on ImageNet~\cite{imagenet} + Kinetics-400~\cite{kinetics}.

\item \textbf{Pace Prediction~\cite{pace}.} This method learns video representations by predicting pace of the video clips in self-supervised manner. They consider five pace candidates that are super slow, slow, normal, fast, super fast.

\item \textbf{Clip Order Prediction~\cite{COP}.} This method learns spatiotemporal representations by predicting the order of shuffled clips from the video in a self-supervised manner.

\item \textbf{CoCLR~\cite{CoCLR}.} It is a self-supervised contrastive learning approach which learns visual-only features using multi-modal information like RGB stream and optical flow
    
\end{itemize}

% \paragraph{\textbf{Implementation Details.}}
% For training pace prediction method, we follow the authors code~\cite{pace} to train the I3D backbone in self-supervised manner. We train the I3D model for 100 epochs with an initial learning rate of 0.001, and a batch size of 48. We decay the learning rate by 0.1 at 35$^{th}$ and 75$^{th}$ epoch. For clip order prediction, we train the I3D for 800 epochs with a batch size of 16 and a learning rate of 0.001. For our approach, we train the model for 350 epochs with a batch size of 24 and a learning rate of 0.1. The temperature parameter $\tau$ is set to 0.1 and the Sinkhorn regularization parameter $\epsilon$ is set to 0.05. We use a queue length of 1920 and number of prototypes is set to 3000. Similar to~\cite{swav}, we use a 2-layer MLP on the top of I3D as a projection head, and it projects the output to 128-D space. We train all the models on 4 NVIDIA RTX A4000 GPUs.

\begin{table}[h!]
\centering
\caption {\label{tab:activity_v1} Mean average precision (mAP) \%  for Bi-GRU under different data regime for surgical activity recognition. }

\resizebox{1.0\textwidth}{!}{%
\begin{tabular}{c|c|c|c|c|c|c}
\hline

\hline

\hline
Methods & \multicolumn{1}{c|}{Pre-train} & 5\% Labeled & 10\% Labeled & \begin{tabular}[c]{@{}l@{}}20\% Labeled\end{tabular} & \begin{tabular}[c]{@{}l@{}}50\% Labeled\end{tabular} & 100\% Labeled \\ \hline

Baseline & None  & 36.31 &  55.77 & 75.70  & 86.09 & 90.71 \\ \hline

Baseline & ImageNet + Kinetics-400  & 37.00 & 55.86 & 76.50   &87.58  & 90.45 \\ \hline

Pace Prediction~\cite{pace} & Dataset~\cite{aidean}   & 39.34  &  63.97& 82.69  &91.63  & 91.86\\ \hline

Clip Order Prediction~\cite{COP} & Dataset~\cite{aidean} & 38.03 &  62.65 & 82.44 & 89.34 & 91.76\\ \hline

CoCLR~\cite{CoCLR} & Dataset~\cite{aidean} & - & 64.87 & 83.74 & - & - \\ \hline

Ours &Dataset~\cite{aidean}  &\textbf{40.27}  & \textbf{67.52} & \textbf{85.20} & \textbf{91.13} & \textbf{92.40} \\ \hline

\end{tabular}
}%
%\vspace{-10pt}
\end{table}

\paragraph{\textbf{Results.}} 
Table~\ref{tab:activity_v1} shows the mean average precision of Bi-GRU with I3D backbone under different data regime. It is clear that our approach outperforms the competing ones in each data regime. In general, the advantages of our approach over existing ones become more significant as the number of labeled data
decreases. We outperform the pace prediction and clip order prediction by +3.5\% and +2.5\% on 10\% and 20\% labeled data respectively. Moreover, when we have 100\% labeled data available in training set, our approach still outperforms the baselines despite the fact that our main goal is to have better pre-trained model for low-data regime.

Table~\ref{tab:activity_v2} shows the mean average precision of Bi-GRU with I3D backbone under different data regime. In this setting, I3D backbone is frozen during training. The purpose of this experiment is to check the quality of the representations learned by different approaches. We can draw the same observation that our approach outperforms the competing ones in each data regime.

\begin{table}[t!]
\centering
\caption {\label{tab:activity_v2} Mean average precision (mAP) \%  for Bi-GRU under different data regime for surgical activity recognition when I3D model is frozen during the training. }

\resizebox{1.0\textwidth}{!}{%
\begin{tabular}{c|c|c|c|c|c|c}
\hline

\hline

\hline
Methods & \multicolumn{1}{c|}{Pre-train} & 5\% Labeled & 10\% Labeled & \begin{tabular}[c]{@{}l@{}}20\% Labeled\end{tabular} & \begin{tabular}[c]{@{}l@{}}50\% Labeled\end{tabular} & 100\% Labeled \\ \hline

Baseline & ImageNet + Kinetics-400  & 35.56 & 40.96 & 50.93   &53.32  & 
53.55\\ \hline

Pace Prediction~\cite{pace} & Dataset~\cite{aidean}   & 36.88
  &  44.03 & 54.09  &72.51  & 80.44\\ \hline

Clip Order Prediction~\cite{COP} & Dataset~\cite{aidean} & 36.33 & 41.98  & 50.71 & 55.67 &69.00 \\ \hline

Ours &Dataset~\cite{aidean}  &\textbf{37.63}  & \textbf{43.43} & \textbf{57.19} & \textbf{73.27} & \textbf{80.47} \\ \hline

\end{tabular}
}%
%\vspace{-5pt}
\end{table}

%& Backbone + Temporal Module

\begin{table}[h!]
\centering
\caption {\label{tab:proto} We report the mean average precision (mAP) \% on low-data regime for surgical activity recognition by varying the number of \textbf{K}.}

\resizebox{0.4\textwidth}{!}{%
\begin{tabular}{c|ccc}
\hline

\hline

\hline
Labeled Data & \multicolumn{3}{c}{Number of prototypes}                     \\ \hline
             & \multicolumn{1}{c|}{1000} & \multicolumn{1}{c|}{2000} & \multicolumn{1}{c}{3000} \\ \hline
5\%          & \multicolumn{1}{c|}{40.39}     & \multicolumn{1}{c|}{40.48}     & \multicolumn{1}{c}{40.27}     \\ \hline
10\%          & \multicolumn{1}{c|}{66.84}     & \multicolumn{1}{c|}{67.40}     &  \multicolumn{1}{c}{67.52}   \\ \hline

\end{tabular}
}%
\vspace{-10pt}
\end{table}

\paragraph{\textbf{Influence of number of prototypes K.}}
We study the impact of the number of prototypes \textbf{K}. In Table~\ref{tab:proto}, we report the mAP \% on low-data regime for surgical activity recognition by varying the number of prototypes. We don't see significant impact on the performance when using 5\% labeled data. For 10\% labelled data, we see a slight drop in the performance for \textbf{K} = 1000. Our work shares the same spirit of~\cite{swav} which also observe no significant impact for ImageNet by varying the number of \textbf{K}.

\subsection{Semantic Segmentation}
We use deeplab-v2~\cite{deeplab} with ResNet-50 as a backbone architecture for this experiment. We only train the ResNet-50 during pre-training step for all the competing methods and our approach. Please refer to supplementary material for implementation details.

\paragraph{\textbf{Competing methods.}}
We compare our approach to the following clustering-based competing baselines.
\begin{itemize}

\item \textbf{Baseline.} This is the baseline that trains the deeplabv2 from the scratch.

\item \textbf{SELA~\cite{sela}.} This method trains a model to learn representations based on clustering by maximizing the information between the input data samples and labels. The labels are obtained using self-labeling method that cluster the samples into K distinct classes.

\item \textbf{Deep Cluster~\cite{deepcluster}.} It is a clustering based approach to learn visual features. It employs k-means for cluster assignments, and subsequently uses these assignments as supervision to train the model.

\end{itemize}

% \paragraph{\textbf{Implementation Details.}}
% For SELA, We follow authors code~\cite{sela}d to train the backbone with a batch size of 256 and an initial learning rate of 0.08 for a total of 250 epochs. The number of clusters is set to 3000. For deep cluster, we follow~\cite{swav} to train the backbone which apply various training improvements that includes multi-clustering, MLP projection head, and cosine learning rate scheduler, temperature parameter to original deep cluster approach. We train the model for 400 epochs with a batch size of 64, and a base learning rate of 4.8. The number of clusters are set 1000 and temperature parameter is set to 0.1. For our approach, we train the model for 75 epochs with a learning rate of 2e-4, and a batch size of 32.  The temperature parameter $\tau$ is set to 0.1 and the Sinkhorn regularization parameter $\epsilon$ is set to 0.03. We use a queue length of 1000 and number of prototypes is set to 50. We train all the models on 4 NVIDIA RTX A4000 GPUs.

\begin{table}[h!]
\centering
\caption {\label{tab:semantic} Mean intersection over union (mIoU) for deeplab-v2 with ResNet-50 backbone under low-data regime for semantic segmentation. }

\resizebox{1.0\textwidth}{!}{%
\begin{tabular}{c|c|c|c|c|c}
\hline

\hline

\hline
Methods & \multicolumn{1}{c|}{Pre-train} & 2\% Labeled & 5\% Labeled & \begin{tabular}[c]{@{}l@{}}10\% Labeled\end{tabular} & \begin{tabular}[c]{@{}l@{}}15\% Labeled\end{tabular} \\ \hline

Baseline & None  & 0.452$\pm$0.008 &  0.483$\pm$0.007 & 0.500$\pm$0.012 & 0.521$\pm$0.010
 \\ \hline

SELA~\cite{sela} & Dataset~\cite{aidean,Schmidt}   & 0.464$\pm$0.009 &  0.499$\pm$0.010 & 0.518$\pm$0.013 & 0.532$\pm$0.006
 \\ \hline

DeepCluster~\cite{deepcluster} & Dataset~\cite{aidean,Schmidt} & 0.481$\pm$0.006 &  0.498$\pm$0.011 & 0.520$\pm$0.004 &0.535$\pm$0.008
 \\ \hline
 
SwAV~\cite{swav} & Dataset~\cite{aidean,Schmidt} & 0.484$\pm$0.006 &  0.502$\pm$0.006 & 0.522$\pm$0.004 & -
 \\ \hline

Ours &Dataset~\cite{aidean,Schmidt}  &\textbf{0.494$\pm$0.014}  & \textbf{0.516$\pm$0.005} & \textbf{0.538$\pm$0.012} & \textbf{0.553$\pm$0.010} \\ \hline

\end{tabular}
}%
%\vspace{-5pt}
\end{table}

\paragraph{\textbf{Results.}}
Table~\ref{tab:semantic} shows the mean intersection over union (mIoU) under low-data regime. For evaluation, we regenerate the subsets and train the model five times to report mean and standard deviation. It is clear that our approach outperforms the competing approaches on low-data regime. This shows that our approach is task-agnostic, and can be used as pre-training step if provided with multi-modal data.

\subsection{Remarks}
While SwAV~\cite{swav} relies on producing difference augmentations (views) of the same image, we propose a multi-modal fusion approach in which different modalities (intensity and depth in our case) are treated as two different views of the same video frame. Moreover, unlike SwAV~\cite{swav}, we show the effectiveness of our approach on video domain. Finally, table~\ref{tab:semantic} shows that our approach still outperforms the SwAV baseline for semantic segmentation. 
% For SwAV baseline, we get mIoU of 0.484±0.006, 0.502±0.006, 0.522±0.004 for 2\%, 5\% and 10\% labeled data which is still lower compared to ours

\section{Conclusion}
In this paper, we propose an unsupervised pretraining approach for video and image analysis task for surgical OR that can enable workflow analysis. Our novel approach combine the intensity and the depth map of a single video frame or image captured from the surgical OR to learn unsupervised representations. While the recent self-supervised or unsupervised learning methods require different augmentations ('views') of a single image, our method considers the intensity and the depth map as two different views. While we demonstrate the effectiveness of our approach on surgical video activity recognition and semantic segmentation in low-data regime, it can also be extended to other tasks where similar multi-modal data is available such as video and image analysis task in laparoscopic surgery. Furthermore, it can be used in pre-training stage for other downstream tasks such as 2D/3D pose estimation, person detection and tracking etc.

%which makes it both model-agnostic and data-agnostic.

%We demonstrate the effectiveness of our approach on surgical video activity recognition and semantic segmentation in low-data regime. We believe that our approach is model-agnostic and can applied to other task where similar multi-modal data is available such as video and image analysis task in endoscopic surgery.
%Our training framework is inspired by recent clustering-based unsupervised approach~\cite{swav} which contrasts the cluster assignments instead of representations.

\bibliographystyle{splncs04}
\bibliography{main}

\newpage
\appendix
\section*{Appendices}
\addcontentsline{toc}{section}{Appendices}
\renewcommand{\thesubsection}{\Alph{subsection}}

\section{Implementation Details.}

\paragraph{\textbf{Surgical Video Activity Recognition.}}
For training pace prediction method, we follow the authors code~\cite{pace} to train the I3D backbone in self-supervised manner. We train the I3D model for 100 epochs with an initial learning rate of 0.001, and a batch size of 48. We decay the learning rate by 0.1 at 35$^{th}$ and 75$^{th}$ epoch. For clip order prediction, we train the I3D for 800 epochs with a batch size of 16 and a learning rate of 0.001. For our approach, we train the model for 350 epochs with a batch size of 24 and a learning rate of 0.1. The temperature parameter $\tau$ is set to 0.1 and the Sinkhorn regularization parameter $\epsilon$ is set to 0.05. We use a queue length of 1920 and number of prototypes is set to 3000. Similar to~\cite{swav}, we use a 2-layer MLP on the top of I3D as a projection head, and it projects the output to 128-D space. We train all the models on 4 NVIDIA RTX A4000 GPUs.

\paragraph{\textbf{Semantic Segmentation.}}
For SELA, We follow authors code~\cite{sela} to train the backbone with a batch size of 256 and an initial learning rate of 0.08 for a total of 250 epochs. The number of clusters is set to 3000. For deep cluster, we follow~\cite{swav} to train the backbone which apply various training improvements that includes multi-clustering, MLP projection head, and cosine learning rate scheduler, temperature parameter to original deep cluster approach. We train the model for 400 epochs with a batch size of 64, and a base learning rate of 4.8. The number of clusters are set 1000 and temperature parameter is set to 0.1. For our approach, we train the model for 75 epochs with a learning rate of 2e-4, and a batch size of 32.  The temperature parameter $\tau$ is set to 0.1 and the Sinkhorn regularization parameter $\epsilon$ is set to 0.03. We use a queue length of 1000 and number of prototypes is set to 50. We train all the models on 4 NVIDIA RTX A4000 GPUs.

% \section{Influence of number of prototypes K.}
% We study the impact of the number of prototypes \textbf{K}. In Table~\ref{tab:proto}, we report the mAP \% on low-data regime for surgical activity recognition by varying the number of prototypes. We don't see significant impact on the performance when using 5\% labeled data. For 10\% labelled data, we see a slight drop in the performance for \textbf{K} = 1000. Our work shares the same spirit of~\cite{swav} which also observe no significant impact for ImageNet by varying the number of \textbf{K}.

% \begin{table}[t]
% \centering
% \caption {\label{tab:proto} We report the mean average precision (mAP) \% on low-data regime for surgical activity recognition by varying the number of \textbf{K}.}

% \resizebox{0.5\textwidth}{!}{%
% \begin{tabular}{c|ccc}
% \hline

% \hline

% \hline
% Labeled Data & \multicolumn{3}{c}{Number of prototypes}                     \\ \hline
%              & \multicolumn{1}{c|}{1000} & \multicolumn{1}{c|}{2000} & \multicolumn{1}{c}{3000} \\ \hline
% 5\%          & \multicolumn{1}{c|}{40.39}     & \multicolumn{1}{c|}{40.48}     & \multicolumn{1}{c}{40.27}     \\ \hline
% 10\%          & \multicolumn{1}{c|}{66.84}     & \multicolumn{1}{c|}{67.40}     &  \multicolumn{1}{c}{67.52}   \\ \hline

% \end{tabular}
% }%
% \end{table}

\end{document}